\documentclass[letterpaper, 10 pt, conference]{ieeeconf}  % Comment this line out if you need a4paper

\IEEEoverridecommandlockouts                              % This command is only needed if 
\usepackage{graphics} % for pdf, bitmapped graphics files
\usepackage{epsfig} % for postscript graphics files
\usepackage{mathptmx} % assumes new font selection scheme installed
\usepackage{times} % assumes new font selection scheme installed
\usepackage{amsmath} % assumes amsmath package installed
\usepackage{amssymb}  % assumes amsmath package installed
\usepackage{pifont}
\usepackage{booktabs}
\usepackage{subcaption}
\usepackage{adjustbox}
\usepackage{array}
\usepackage{multirow, booktabs}
\usepackage{makecell}
\usepackage[mathscr]{euscript}
\usepackage{float}
\usepackage{gensymb}
\usepackage[table,x11names]{xcolor} % for \rowcolor and specific colors
\definecolor{PaleBlue}{rgb}{0.87,0.92,1}
\definecolor{PaleGreen}{rgb}{0.87, 1.0, 0.92}
\usepackage{mathtools}
\usepackage{todonotes}
\usepackage{booktabs}
\let\labelindent\relax
\usepackage{enumitem}

\usepackage{algpseudocode}
\usepackage{cite}
\usepackage{booktabs}
\usepackage{listings}
\usepackage[linesnumbered,algoruled,boxed,vlined, noend]{algorithm2e}

\title{\LARGE \bf
DAP: Diffusion-based Affordance Prediction for Multi-modality Storage
}

\author{Haonan Chang, Kowndinya Boyalakuntla, Yuhan Liu, Xinyu Zhang, Liam Schramm, Abdeslam Boularias% <-this % stops a space
\thanks{The authors are with the Department of Computer Science,
        Rutgers University, 08854 New Brunswick, USA. This work is supported by NSF awards 1846043 and 2132972.}}
\begin{document}

\maketitle
\thispagestyle{empty}
\pagestyle{empty}

%%%%%%%%%%%%%%%%%%%%%%%%%%%%%%%%%%%%%%%%%%%%%%%%%%%%%%%%%%%%%%%%%%%%%%%%%%%%%%%%

\begin{abstract}
Solving storage problems—where objects must be accurately placed into containers with precise orientations and positions—presents a distinct challenge that extends beyond traditional rearrangement tasks. These challenges are primarily due to the need for fine-grained 6D manipulation and the inherent multi-modality of solution spaces, where multiple viable goal configurations exist for the same storage container. We present a novel Diffusion-based Affordance Prediction (DAP) pipeline for the multi-modal object storage problem. DAP leverages a two-step approach, initially identifying a placeable region on the container and then precisely computing the relative pose between the object and that region. Existing methods either struggle with multi-modality issues or computation-intensive training. 
Our experiments demonstrate DAP's superior performance and training efficiency over the current state-of-the-art RPDiff, achieving remarkable results on the RPDiff benchmark. Additionally, our experiments showcase DAP's data efficiency in real-world applications, an advancement over existing simulation-driven approaches. Our contribution fills a gap in robotic manipulation research by offering a solution that is both computationally efficient and capable of handling real-world variability. Code and supplementary material can be found at: {\small https://github.com/changhaonan/DPS.git}.

\end{abstract}

\begin{figure*}[htp]
    \centering
    \includegraphics[width=1.0\linewidth]{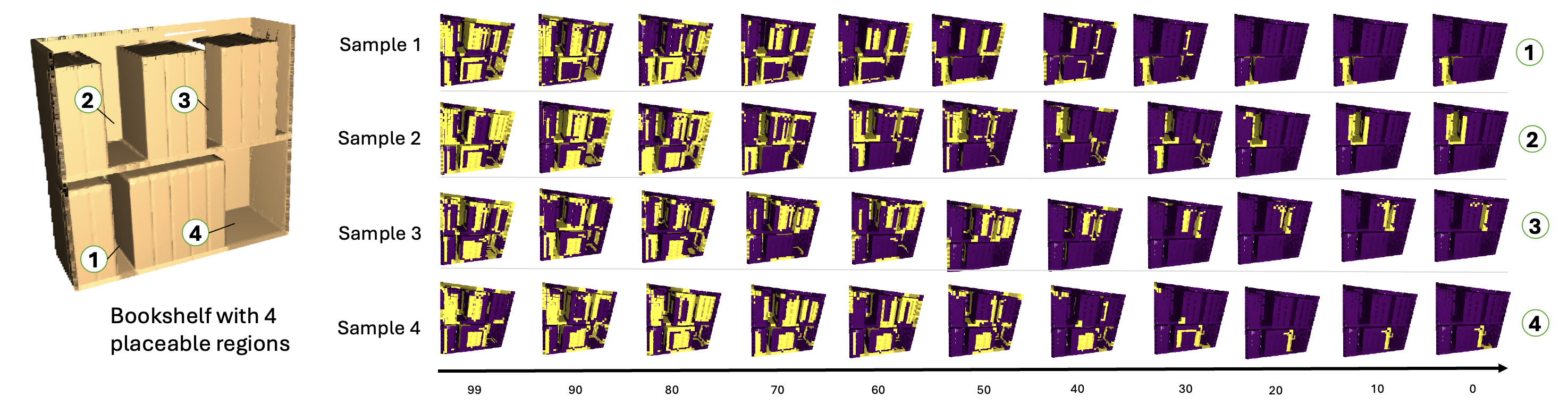}
    \caption{\small{Visualization of the backward diffusion process in affordance prediction. Each row represents different samples. Each column corresponds to one diffusion step. The diffusion step $t$ starts from 99 and ends at 0. Each figure represents a visualization of a sample at that time step. Yellow indicates that the region is placeable, and purple indicates it is not. At beginning, the scene starts with a random segmentation. As the backward diffusion process progresses, the affordance prediction gradually converges to the 4 placeable regions. }}
    \label{fig:afford_illu}
    \vspace{-20px}
\end{figure*}

\section{INTRODUCTION}
Storage tasks, such as placing a plate into a dishwasher or putting a book onto a bookshelf, are ubiquitous in our daily lives. These tasks involve placing an object into a container, with the pose of the placed object meeting specified criteria. However, unlike general rearrangement problems, storage problems present two unique challenges: strict geometrical constraints and multi-modal solutions. Firstly, the storage criteria necessitate either an in-contact or a near-contact goal pose configuration that is physically stable, such as the case when inserting a book vertically into a tight gap on a bookshelf. Furthermore, the entire placing process must be collision-free. Secondly, there typically exist multiple functionally correct but geometrically different goal configurations under the same storage criterion. This inherent multi-modality significantly impacts regression-based models, such as Coarse-to-fine Q-attention~\cite{james2022coarse}, Relational Neural Descriptor Fields~\cite{se3}, Neural Shape Mating~\cite{chen2022neural}, or Structformer~\cite{liu2022structformer}. 

Diffusion models have been shown to address the multi-modality issue in image~\cite{ddpm} and video generation. StructDiffusion~\cite{liu2022structdiffusion} pioneered the use of diffusion models in rearrangement tasks by using diffusion to model the distribution of task scenes. However, it suffers from inaccurate pose prediction. Relative Pose Diffusion (RPDiff) proposes {\it Pose-diffusion}~\cite{shelving}, where initially-random relative poses are iteratively refined by a denoising model until an accurate goal pose is found. However, RPDiff requires a significant amount of environmental interactions as training data, making it viable only in simulated tasks and not in real robotic tasks.

In this work, we introduce the \textbf{D}iffusion-based \textbf{A}ffordance \textbf{P}rediction (DAP) method to address storage problems. Our key insight is to disentangle the strict geometrical constraint and the multi-modality issue by tackling them separately. Rather than directly predicting a goal pose within the entire scene, our method, DAP, initially identifies a placeable region within only the container region through diffusion-based affordance prediction. Unlike classical affordance prediction, which locates all placeable regions, DAP models the multi-modal distribution of placeable regions using a diffusion model. Next, inspired by region matching in one-shot manipulation learning~\cite{zhang2024oneshot}, DAP derives the goal pose by finding a point-wise correspondence between the object and the identified region, without the interference from other possible placeable regions within the container. For example, when placing a plate into a dishwasher, DAP learns to model the distribution of valid slots, samples one slot, and then deterministically solves for the goal pose of that slot. Our experiments demonstrate that DAP effectively resolves the multi-modality issue while predicting accurate goal poses. Compared to RPDiff, our method can be trained in just 2 hours, instead of several days.

Our contributions are summarized as follows: (1) We propose DAP, an efficient diffusion-based method that predicts accurate goal poses for storage problems by generating a multi-modal affordance distribution. (2) We evaluate DAP on the RPDiff simulated benchmark, demonstrating that our method is significantly more training-efficient and achieves better accuracy compared to the existing state-of-the-art, RPDiff~\cite{shelving}.  (3) We deploy DAP in a real-robot system, where it is shown to perform real-world storage tasks effectively, even with noisy observations and minimal training data.

% In this work, we introduce the \textbf{D}iffusion-based \textbf{A}ffordance \textbf{P}rediction (DAP) method to solve storage problems. Our key insight is to disentangle the strict geometrical constraint and the multi-modality issue by solving them separately. 
% Instead of predicting a goal pose directly, our method DAP first identifies a placeable region inside the container through a diffusion-based affordance prediction. 
% Unlike classical affordance prediction which locates all placeable regions, DAP models the multi-modal affordance distribution over placeable regions using a diffusion model. 
% Next, DAP derives the goal pose by finding correspondence between the object and the identified region, without the interference of other possible placeable regions in the container. 
% For example, when placing a plate into a dishwasher, DAP learns to model the distribution of valid slots, sample one slot, and then deterministically solve the goal pose for that slot. 
% Our empirical evaluation shows that DAP solves the multi-modality issue while predicting accurate goal poses. 
% Compared to RPDiff, our method is applied to real-world storage tasks and can be trained in 2 hours instead of several days. 
\section{RELATED WORKS}
\label{related}
\subsection{Pair-wise Object Manipulation}
Storage requires to compute the precise transformation between the object being moved and the stationary container object. In the realm of pair-wise object manipulation, traditional approaches start with point cloud registration to identify the task-relevant region, followed by relative transformation estimations. Tax-Pose~\cite{tax_pose} and R-NDF~\cite{se3} exemplify this, using transformers and neural descriptor fields to compute the correspondence between the stationary object and the moving object and infer transformations, respectively. On the other hand, Neural Shape Mating~\cite{chen2022neural} learns the transformation directly, without the point cloud registration. However, these methods struggle with multi-modal tasks, a gap filled by RPDiff's~\cite{shelving} diffusion-based pose refinement model, at a high computational cost requiring several days of training on an advanced GPU (V100). Our approach presents a solution to the computational and multi-modality challenges inherent in the existing works. 
We achieve this by proposing a diffusion-based affordance prediction method to retrieve one task-relevant region among many and then determine the correspondences between the identified sub-region and the moving object to estimate the transformation accurately.

\subsection{Affordance Prediction  \& Point Cloud Segmentation}
In 3D point cloud segmentation, methods are classified into: (1) Semantic segmentation, categorizing points into broad classes; (2) Instance segmentation, identifying individual entities; and (3) Affordance segmentation, segmenting object regions for interactions like pushing or storing. Initially, the focus was on MLP/CNN-based architectures (e.g., PointNet~\cite{pointnet}, PointNet++~\cite{pointnet++}, 3DSIS~\cite{3dsis}, PointGroup~\cite{pointgroup}), but recent advances have shifted towards transformer-based models (e.g., Superpoint~\cite{sun2023superpoint}, Point Transformer~\cite{point_transformer}, Mask3D~\cite{mask3d}, OneFormer3D~\cite{oneformer3d}), starting with Point Transformer~\cite{point_transformer}'s advancement in semantic segmentation. Subsequent works like Mask3D~\cite{mask3d} and OneFormer3D~\cite{oneformer3d} have pushed the boundaries in semantic and instance segmentation with transformer models, enhancing performance. In affordance prediction on 3D point clouds~\cite{3daffordancenet, 3dapnet, o2o, openvocab_affordance, openvocab_knowledge}, 3D AffordanceNet~\cite{3daffordancenet} provides a benchmark across 18 affordance categories, with 3DAPNet~\cite{3dapnet} simultaneously predicting affordance regions and generating corresponding 6DoF poses for action affordances.

Our problem is to segment a suitable region among many possible ones for object storage, introducing complexities beyond the scope of traditional segmentation approaches. Semantic segmentation cannot distinguish between multiple viable regions. Instance segmentation is impractical due to the variability of potential storage spaces (e.g., placing a can on a cabinet with many stackable regions), making the generation of instance labels infeasible. Our task aligns more with affordance prediction, aiming to segment a region from a container object. However, existing methods do not address multi-modality, failing to select a single storage region from multiple candidates. To this end, we combine the diffusion model with the latest Point Transformer architecture to model the distribution of placeable storage region within a container's point cloud. 
\subsection{Diffusion Model}

Diffusion models have seen success in a wide range of generative tasks, including image generation \cite{dhariwal2021diffusion, ddpm}, video generation \cite{liu2024sora}, imitation learning \cite{chi2023diffusionpolicy}, and offline reinforcement learning \cite{wang2023diffusion}. These models are latent variable models that consist of two processes: a noising \textit{forward process}, in which Gaussian noise is iteratively added to data samples, and a denoising \textit{backward process} in which a learned model predicts what noise was added in the forward process and removes it to reconstruct the original data sample. The model is trained by minimizing the mean-squared error between the predicted noise and the actual noise \cite{ddpm}. This method offers several benefits over other generative architectures. Compared to GANs, diffusion models are more stable during training because they do not involve solving a minimax problem \cite{mescheder2018training, kodali2017convergence}. Compared to approximating the target distribution with a multivariate Gaussian, diffusion models can represent arbitrary probability distributions, so they perform better in settings where it is important to represent multi-modality \cite{debortoli2023convergence}.

Two major milestones in the development of diffusion models are Deep Denoising Probabilistic Models (DDPM) ~\cite{ddpm} and Diffusion Transformers (DiT) ~\cite{diffusion_transformer}. DDPM uses the Rao-Blackwell theorem to obtain a closed-form expression for the noise target, which makes training considerably faster. DiT uses neural networks with a transformer architecture, enabling better scaling and generalization to variable-length inputs. We use the diffusion transformer architecture with the DDPM loss function.

\section{Problem formulation}

We address the challenge of multi-modality storage. Our objective is to position a target object $O$ inside a bigger container $C$, considering that there are multiple viable placements for $O$ within $C$. We represent the relative transformation between $O$ and $C$ as $\mathbf{T}_{OC}\in \mathbb{SE}(3)$. The storage is successful when $\mathbf{T}_{OC}$ falls in the support of a multi-modal distribution $\mathcal{D}$. The goal is to, given the point cloud observations of $O$ and $C$, $\mathbf{P}_O$ and $\mathbf{P}_C$, in the world coordinate system $W$, calculate a transformation for $O$, denoted as $\mathit{\mathbf{T}_{WO}}=(\mathit{\mathbf{R}_{WO}} \in \mathbb{SO}(3), \mathit{\mathbf{t}_{WO}}\in \mathbb{R}^3)$. Applying this transformation to object $O$ should result in the relative pose of $O$ and $C$ falling into the distribution $\mathcal{D}$. Point cloud $\mathbf{P}$ consists of point vertexes $\{v_i\}_{i=1}^N$ and normals $\{n_i\}_{i=1}^N$. We assume a small set of $M$ demonstrations $\{\mathbf{P}_O^j, \mathbf{P}_C^j, \mathbf{T}_{WO}^j\}_{j=1}^M$ is provided.
\section{Method}

% \subsection{Overview}
We tackle this problem using a two-stage method. Initially, we employ a diffusion-based affordance prediction to identify the placeable regions within the container, given the target object. Unlike conventional affordance prediction methods, which return all placeable regions simultaneously without distinction, our diffusion-based approach singles out one focused region in each sample. Upon identifying the placeable region, we proceed to compute the relative pose between the placeable region and the target object. Rather than directly calculating the $\mathbb{SE}(3)$ transformation, we first establish a point-wise correspondence between the container's local region and the target object's point cloud. This correspondence predicts which parts of the container and target should be in contact. We then utilize the algorithm in~\cite{arun1987least} to determine the pose from this correspondence.

\subsection{Diffusion-based Affordance Prediction}
\label{sec:dap}
The primary challenges in the multi-modal storage problem are twofold:  (1) The model must have high enough accuracy that the generated poses are stable and avoid collisions, and (2) The multi-modal nature of the task presents multiple viable solutions, making it difficult for learning-based methods to separate them. To address the first issue, we adopt a coarse-to-fine strategy, proven by prior research~\cite{james2022coarse} to enhance pose prediction accuracy effectively. In tackling the second challenge of ambiguity of viable solutions, we introduce a diffusion-based affordance prediction method. This method serves as a critical step in our coarse-to-fine strategy, effectively narrowing down the possibilities by focusing on placeable regions within the container.  Specifically, we aim to predict a score $\mathbf{S} =(s_1,s_2,\dots,s_{N_C}), s_i \in [-1, 1]$ for each point in the container point cloud $\mathbf{P}_C$, where a higher score signifies a more suitable placement area. After we obtain the affordance prediction $\mathbf{S}$, we crop the container based on this prediction, and then perform pose-relevant computation on that local geometry. This prediction is framed as a generative task, aiming to model the conditional distribution of score $\mathbf{S}$ over container geometry $\mathbf{P}_C$. 

\noindent \textbf{Data labeling:} 
As outlined in the problem formulation, our data comprises $\{\mathbf{P}_C,\mathbf{P}_O,\mathbf{T}_{WO}\}$. From this, we need to generate labels for placeable affordance. We apply the transformation $\mathbf{T}_{WO}=(\mathbf{R}_{WO}, \mathbf{t}_{WO})$ to $\mathbf{P}_O$ using the formula:
\begin{align}
    v_i' = \mathbf{R}_{WO} v_i + \mathbf{t}_{WO},\ n_i' = \mathbf{R}_{WO} n_i.
    \label{eq:transform}
\end{align}
This results in the transformed point cloud $\mathbf{P}_O'=\{(v_i', n_i')\}_{i=1}^{N_O}$, which represents the goal object point cloud. Next, we identify points on the container $\mathbf{P}_C$ whose minimal distance to the transformed object $\mathbf{P}_O'$ is smaller than a threshold $\epsilon_{place}$. These nearby points to the target point cloud on the container point cloud indicate the placeable region on $\mathbf{P}_C$. We assign a score of $1$ to these points and $-1$ to the rest. Formally, this labeling is defined as:
\begin{equation}
    s_i = \begin{cases} 1 & \text{if\ }\underset{v_j \in \mathbf{P}_C}{\min}||v_i' - v_j||_2 < \epsilon_{place},\ v_i' \in \mathbf{P}_O'\\
    -1 & \text{else}
    \end{cases}
\end{equation}
    
Here, $\epsilon_{place}$ serves as a hyper-parameter to adjust the size of the placeable region, enabling us to mitigate the ambiguity inherent in the multi-modality storage challenge.

\noindent \textbf{Training:} 
Based on our label generation method, the score $\mathbf{S}$ will be a distribution conditioned on the container's geometry $\mathbf{P}_C$, denoted as $\mathcal{D}_S = p(\mathbf{S}|\mathbf{P}_C)$. To capture $\mathcal{D}_S$, we utilize a denoising diffusion probabilistic model (DDPM)~\cite{ddpm}. We construct a continuous diffusion process $\{\mathbf{S}(t)\}_{t=0}^T$ indexed by time-variable $t$. $\mathbf{S}(0)$ originates from the demonstration data, representing the ground-truth affordance score. As the time-step $t$ progresses from 1 to $T$ (the total number of diffusion steps), $\mathbf{S}(t)$ is progressively perturbed by Gaussian noise,
\begin{equation}
    p(\mathbf{S}(t)|\mathbf{S}(t-1),\mathbf{P}_C) := \mathcal{N}(\mathbf{S}(t); \sqrt{1-\beta_t}\mathbf{S}(t-1),\beta_t I).
\end{equation}
Here $\beta_t$ follows the notation in \cite{ddpm}. The training goal is to learn a network $\mathbf{\mu}_{\theta}(\mathbf{S}(t),t,\mathbf{P}_C)$, which is able to backward the diffusion process, estimating $\mathbf{S}(t-1)$ from $\mathbf{S}(t)$:
\begin{equation}
    p_{\theta}(\mathbf{S}(t-1)|\mathbf{S}(t),\mathbf{P}_C) := \mathcal{N}(\mathbf{S}(t-1); \mathbf{\mu}_{\theta}(\mathbf{S}(t-1), t, \mathbf{P}_C),\sigma_t).
\end{equation}
According to \cite{ddpm}, rather than directly estimating $\mathbf{\mu}_{\theta}$, we can express $\mathbf{\mu}_{\theta}$ as:
\begin{equation}
\mathbf{\mu}_{\theta}(\mathbf{S}(t),t,\mathbf{P}_C) = \frac{1}{\sqrt{\alpha}} (\mathbf{S}(t) - \frac{1 - \alpha_t}{\sqrt{1-\bar{\alpha}_t}}\mathbf{\epsilon}_{\theta}(\mathbf{S}(t),t,\mathbf{P}_C).
\end{equation}
Thus, the training objective for the DDPM can be simplified to:
\begin{equation}
    L_t^\text{simple}
= \mathbb{E}_{t \sim [1, T], \mathbf{S}_0, \boldsymbol{\epsilon}_t} \Big[\|\boldsymbol{\epsilon}_t - \boldsymbol{\epsilon}_\theta(\mathbf{S}(t), t, \mathbf{P}_C)\|^2 \Big].
\end{equation}
The parameters $\alpha_t, \bar{\alpha}_t, \beta_t, \epsilon_t$ adhere to the definitions provided in \cite{ddpm}. This training objective is equal to minimizing a variational lower bound over the KL-divergence between a learned distribution $\mathcal{D}_{\theta}$ and the goal distribution $\mathcal{D}_S$. After training, we can sample from the learned distribution $\mathcal{D}_{\theta}$ with the learned network $\mathbf{\epsilon}_{\theta}(\mathbf{S}(t),t,\mathbf{P}_C)$. We start from a pure Gaussian noise $\mathbf{S}(T)\sim \mathcal{N}(0, I)$, and then perform the denoising steps from $t=T$ to $t=1$ using:
\begin{equation}
    \mathbf{S}(t-1) = \frac{1}{\sqrt{\alpha_t}} (\mathbf{S}(t) - \frac{1 - \alpha_t}{\sqrt{1-\bar{\alpha}_t}}\mathbf{\epsilon}_{\theta}(\mathbf{S}(t),t,\mathbf{P}_C)) + \sigma_t z.
\end{equation}
Here $z\sim \mathcal{N}(0, I)$ if $t>1$ else 0.  And we select $\sigma_t^2 = \beta_t$. After iterating from $t=T$ to $t=1$, we get an affordance prediction $\mathbf{S}(0)$. Fig.~\ref{fig:afford_illu} provides an illustrative visualization for this sampling process.

\noindent \textbf{Architecture:} 
\begin{figure}
    \centering
    \includegraphics[width=1.0\linewidth]{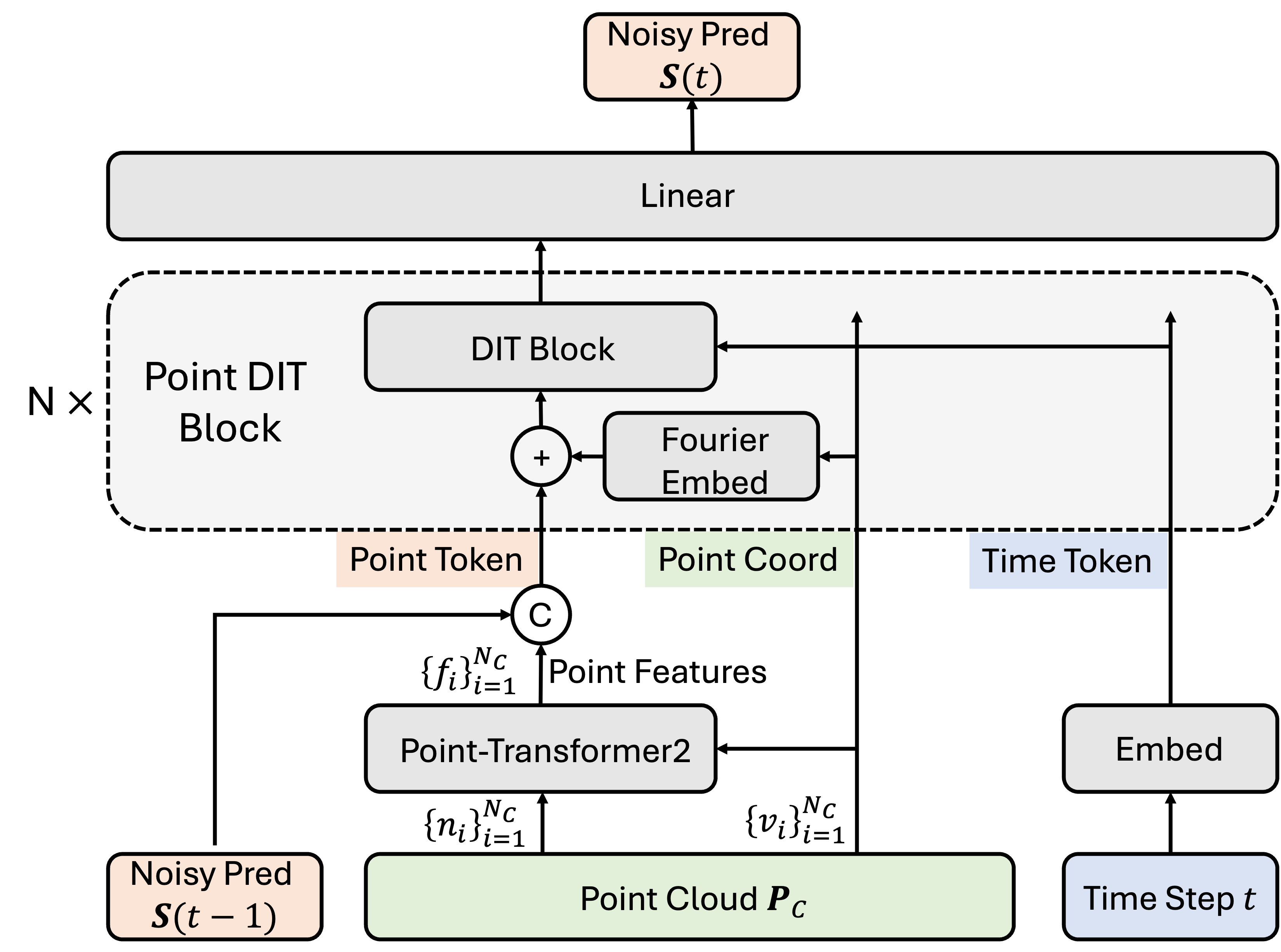}
    \caption{\small{The Diffusion Affordance Prediction Architecture. }}
    \label{fig:dit}
\end{figure}
For the network $\mathbf{\epsilon}_{\theta}(\mathbf{S}(t-1),t, \mathbf{P}_C)$, we adopt a diffusion-transformer (DiT) architecture as introduced in~\cite{diffusion_transformer}. A major distinction is that, whereas the original DiT was designed for 2D tasks, our task is inherently 3D. We detail our architecture in Fig.~\ref{fig:dit}. $\mathbf{\epsilon}_{\theta}(\mathbf{S}(t-1),t, \mathbf{P}_C)$ takes as input the point cloud $\mathbf{P}_C$, the noisy prediction $\mathbf{S}(t-1)$, and the time-step $t$. As illustrated in Fig.~\ref{fig:dit}, the point cloud $\mathbf{P}_C$ is input into the network at two different positions: one part uses the point coordinates $\{v_i\}_{i=1}^{N_C}$, and the other utilizes per-point features $\{f_i\}_{i=1}^{N_C}$. The Point-Transformer2~\cite{wu2022point} serves as the backbone to extract per-point features $\{f_i\}_{i=1}^{N_C}$ from coordinates $\{v_i\}_{i=1}^{N_C}$ and normals $\{n_i\}_{i=1}^{N_C}$. These per-point features $\{f_i\}_{i=1}^{N_C}$ are concatenated with the noisy scores $\{s_i\}_{i=1}^{N_C}$ to form the point tokens. The time-step $t$ is processed through an embedding layer, generating the time token. These point tokens, point coordinates, and the time token are then fed into the Point-DiT block. Within the Point-DiT block, we apply a Fourier position embedding~\cite{li2021learnable} to encode the point-wise positional information. Notably, unlike in traditional transformer architectures where positional encoding is applied only at the first layer, we implement this encoding at every layer. As demonstrated in~\cite{mask3d}, applying positional encoding at each transformer layer proves advantageous for segmentation tasks. Subsequently, the position-encoded point tokens and time-token are processed by the DiT Block, which retains the structure described in~\cite{diffusion_transformer}. The output refined point tokens are then used as input for the next Point-DiT layer, while the point coordinates and time-token remain unchanged. Finally, a linear layer projects the latent embeddings back to an $N_C \times 1$ vector with a range of $[-1, 1]$. 

After obtaining the final affordance prediction $\mathbf{S}$, we crop point cloud $\mathbf{P_C}$ by removing all points with negative scores. %The multi-modality issue is generally addressed by this cropping. %Although within the remaining local region of the container, there exists one optimal solution. For notation simplicity, 
We use $\mathbf{P}^*_C$ to denote the cropped point cloud in Section~\ref{sec:pose_estimation}.

\subsection{Pose estimation}
\label{sec:pose_estimation}
\begin{figure}
    \centering
    \includegraphics[width=1.0\linewidth]{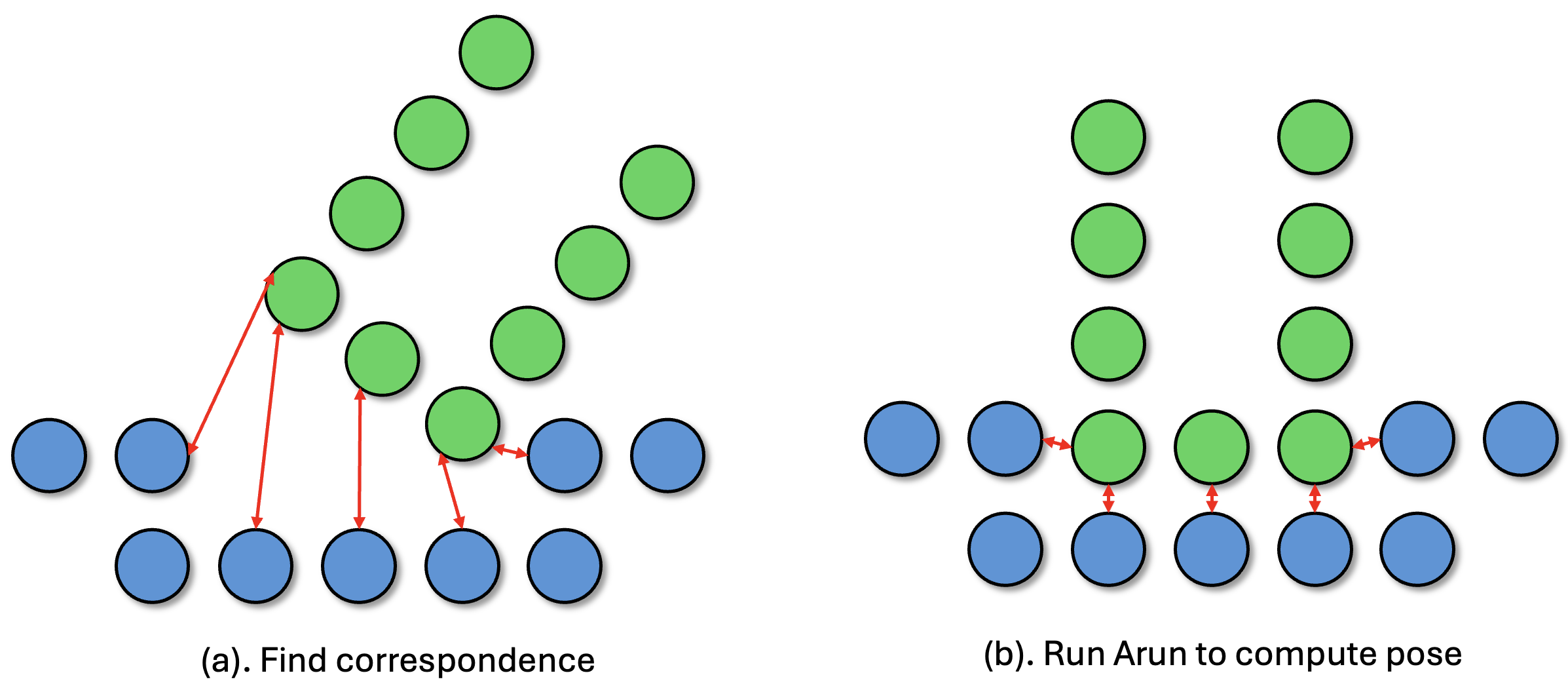}
    \caption{\small{Illustration of the correspondence and pose computation on a 2D toy example. The green points are the target object, and the blue points are the container. }}
    \label{fig:corr_illu}
\end{figure}

As aforementioned, one challenge of multi-modality storage is that it requires high accuracy for the generated placement pose. This is especially crucial for compact regions, such as placing a book on a shelf, where the gap for placing an object may be very small, and we need to ensure the pose we generate is physically plausible and collision-free. Previous works for pairwise object manipulation~\cite{tax_pose,se3} have shown that, rather than directly predicting the pose for an object, decomposing the pose estimation into first finding point-wise correspondence between point cloud and then computing the 6D pose from correspondence seems to be more stable and accurate. We therefore utilize a similar pipeline in our method. We train a network $\mathbf{C}_{\phi}(\mathbf{P}^*_C, \mathbf{P}_O)$ to predict the correspondence matrix $\mathbf{C}$ between two geometries $\mathbf{P}^*_C$ and $\mathbf{P}_O$. This correspondence $\mathbf{C}$ models which point on $\mathbf{P}^*_C$ should be in contact with which point on $\mathbf{P}_O$. Then, we apply Arun’s algorithm, which is a least squares optimization method that minimizes the distance between the corresponding points. Arun's algorithm returns the goal pose, $\mathbf{T}_{WO}$. We present a toy 2D example in Fig.~\ref{fig:corr_illu} to illustrate how our pose estimation pipeline looks like.

\noindent \textbf{Data labeling:} 
We sample a random size bounding box around the demonstrated storage location. We crop $\mathbf{P}_C$ using this bounding box to get $\mathbf{P}^*_C$.
The ground-truth correspondence identifies which parts of $\mathbf{P}^*_C$ and $\mathbf{P}_O$ should be in contact. We apply $\mathbf{T}_{WO}$ to $\mathbf{P}_O$ using  Eq.~\ref{eq:transform}, resulting in $\mathbf{P}_O'$. Subsequently, we calculate the pairwise distance between all points in $\mathbf{P}_O’$ and all points in $\mathbf{P}^*_C$. For any two points in $\mathbf{P}^*_C$ and $\mathbf{P}_O$, their correspondence value is set to 1 if their distance is less than a threshold $\epsilon_{corr}$, and 0 otherwise. Correspondence matrix $\mathbf{C}$'s shape is $(N_O \times N_C)$. Mathematically, $\mathbf{C}$ is defined as follows:
\begin{equation}
\mathbf{C}(i,j) = \begin{cases} 1, & ||v_i' – v_j||_2 < \epsilon_{corr}, v_i' \in \mathbf{P}_O’, v_j \in \mathbf{P}^*_C \\
0, & else
\end{cases}
\end{equation}

% As we mentioned in the end of Section~\ref{sec:dap}, $\math{P}_C$

\noindent \textbf{Training:} The training for correspondence is conducted through pure supervised learning. We assume that the multi-modality problem has been addressed by the diffusion-based affordance prediction, leading to the existence of only one optimal correspondence for given $\mathbf{P}^*_C$ and $\mathbf{P}_O$. To this end, we train a network $\mathbf{C}_{\phi}(\mathbf{P}^*_C, \mathbf{P}_O)$ to approximate $\mathbf{C}$. We employ a focal loss between $\mathbf{C}_{\phi}(\mathbf{P}^*_C, \mathbf{P}_O)$ and the ground-truth $\mathbf{C}$ as training objective:
\begin{equation}
    L^{corr} = \sum_{i=1}^{N_O}\sum_{j=1}^{N_C} \log{\big(\mathbf{C}(i,j)\mathbf{C}_{\phi}(i, j)\big)}\cdot \big(1 - \mathbf{C}(i,j)\mathbf{C}_{\phi}(i, j)\big)^{\gamma}
\end{equation}
The focal loss is specifically chosen to mitigate the imbalance in data distribution. $\gamma$ is a hyper-parameter to tune the balancing strength.

\noindent \textbf{Architecture:} 
\begin{figure}

    \centering
    \includegraphics[width=0.9\linewidth]{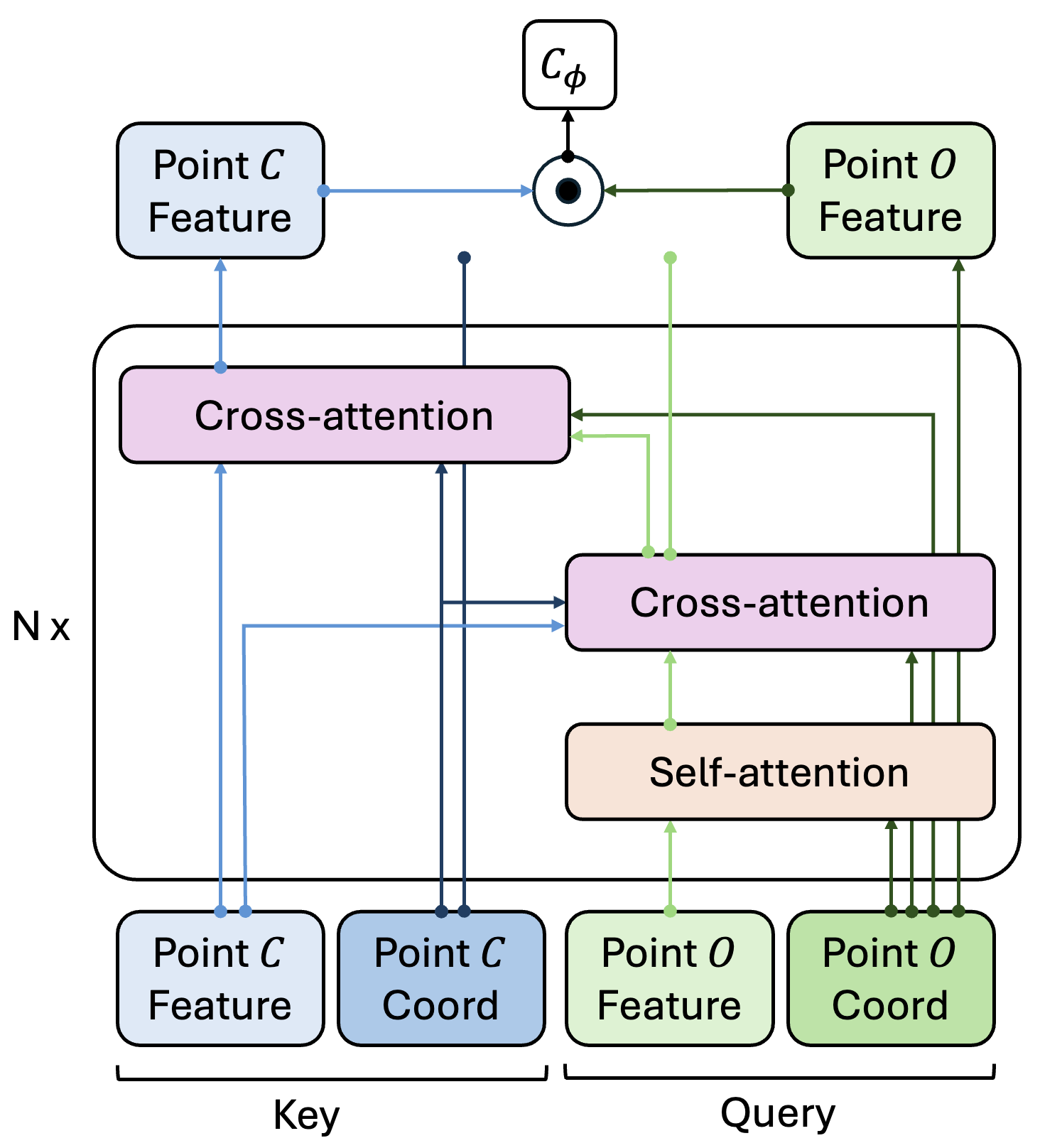}
    \vspace{-0.5em}
    \caption{\small{The correspondence prediction architecture inspired by IMOP~\cite{zhang2024oneshot} }}
    \label{fig:corr}
    \vspace{-2em}
\end{figure}
We employ Point-Transformer2~\cite{wu2022point} as the 3D backbone network to extract point-wise features $\{f_i\}_{i=1}^{N}$ for both $\mathbf{P}^*_C$ and $\mathbf{P}_O$. Point-Transformer2 introduces an efficient attention mechanism termed Grouped Vector Attention (GVA). Unlike classical attention mechanisms that calculate the attention between all key tokens and query tokens, GVA computes attention within predefined groups, necessitating the establishment of these groups beforehand. In 3D problems, where each token is associated with 3D points, we can utilize K-nearest-neighbors (KNN) to determine the attention groups. For instance, to compute a KNN-based GVA between two point clouds $\mathbf{P}_1$ and $\mathbf{P}_2$, where $\mathbf{P}_1$ serves as the query point cloud and $\mathbf{P}_2$ as the key point cloud, we determine the K-nearest neighbors of each point in $\mathbf{P}_1$ within $\mathbf{P}_2$. The attention logit for each point in $\mathbf{P}_1$ is then calculated using this point-token and its K-nearest neighbor point tokens in $\mathbf{P}_2$. Due to space constraints, we refer readers to \cite{wu2022point} for the complete definition of GVA.

In our approach, we use KNN-GVA for efficient self-attention and cross-attention processing on point cloud data. The full correspondence prediction pipeline is depicted in Fig.~\ref{fig:corr}. $\mathbf{P}^*_C$ and $\mathbf{P}_O$ are fed into the backbone point network to extract point-wise features. Object point tokens $\{f_i\}_{i=1}^{N_O}$ act as query tokens, while container point tokens $\{f_i\}_{i=1}^{N_C}$ serve as key tokens. These query tokens are processed through a KNN-GVA layer for self-attention. Subsequently, cross-attention is performed between the query tokens and key tokens to refine the query tokens. This process is followed by cross-attention between key tokens and query tokens to refine the key tokens. The refined query and key tokens are then used as inputs for the next block. Finally, a dot-product operation is employed to predict the correspondence between points:
\begin{equation}
    \mathbf{C}_{\phi}(i, j) = f_i \cdot f_j,\ i \in \mathbf{P}_O, j \in \mathbf{P}^*_C
\end{equation}

\noindent \textbf{Pose solving \& Ranking:} After getting the point-correspondence between $\mathbf{P}_O$ and $\mathbf{P}^*_C$, we can analytically compute the goal pose $\mathbf{T}_{WO}$ using Arun's algorithm~\cite{arun1987least}. 
While the pose estimation step is a deterministic process, the previous step, diffusion-based affordance prediction and cropping, is a sampling process. We sample $K$ candidate poses each time, where $K$ is a hyper-parameter. We perform simple collision checking between the resulting $\mathbf{P}_O'$ and $\mathbf{P}^*_C$: counting how many points of $\mathbf{P}^*_C$ fall within the bounding box of $\mathbf{P}_O'$. Candidates are ranked based on this collision estimation.
\section{EXPERIMENTS}

We conduct a comprehensive series of experiments, encompassing both simulation and real-world scenarios, aiming to address several key questions: (1) How does the performance of DAP compare with other methods? (2) How crucial is the diffusion-based affordance prediction for addressing the multi-modality issue? (3) Is our method sufficiently data-efficient to learn effectively from real-world data?

\subsection{Simulation Experiment} 
\begin{figure}
    \centering
    \includegraphics[width=1.0\linewidth]{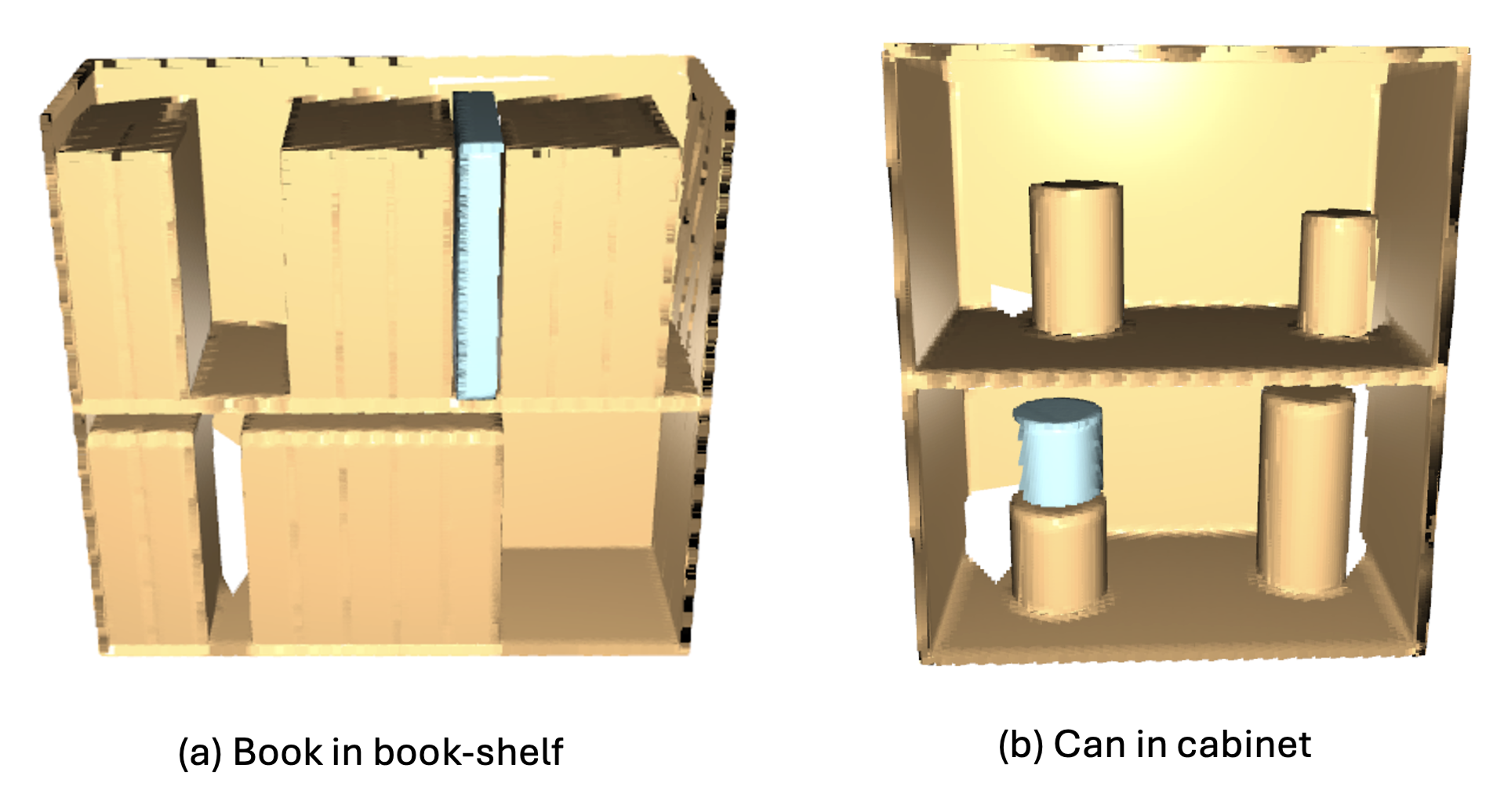}
    \caption{\small{Samples from RPdiff benchmark. We show two sample scenes from the RPdiff benchmark: one is placing a book into the bookshelf and the other is stacking a can inside a cabinet. }}
    \label{fig:rpdiff_data}
    \vspace{-1em}
\end{figure}
We evaluate our method using the benchmark from RPDiff~\cite{shelving} (check Fig.~\ref{fig:rpdiff_data}), which provides a challenging simulation environment for addressing the multi-modal rearrangement problem. This environment includes tasks such as book shelving, can stacking, and cup hanging, all of which highlight the benchmark's complexity due to the variability in container and object geometries within each task. This variability demands a model's ability to generalize across different geometric configurations. We exclude the cup hanging task from evaluation as it does not match our problem requirement that the object is to be placed in a bigger container.
The benchmark's inputs are a container point cloud $\mathbf{P}_C$ and an object point cloud $\mathbf{P}_O$. Success is determined by the object's stable placement inside the container. The reported success rate is averaged over 100 independent random trials.

Regarding the baselines, we adopt the same benchmarks used in RPDiff~\cite{shelving}. We compare our approach against five baseline methods, each offering a unique perspective on tackling multi-modal rearrangement problems:

\noindent \textbf{Coarse-to-Fine Q-attention (C2F-QA):} Adapted from a classification approach, this method predicts a score distribution over a voxelized scene representation to identify candidate translations of the object centroid. It operates in a coarse-to-fine manner, refining predictions at higher resolutions and culminating in a rotation prediction for the object. The best-scoring transformation is then executed.

\noindent \textbf{Relational Neural Descriptor Fields (R-NDF):} Utilizing a neural field shape representation, R-NDF matches local coordinate frames to category-level 3D models, facilitating relational rearrangement tasks. The ``R-NDF-base" version does not include the refinement energy-based model found in the original implementation.

\noindent \textbf{Neural Shape Mating (NSM) + CVAE:} NSM processes paired point clouds via a Transformer to align them. The ``NSM-base" differs in its training on large perturbations without local cropping and makes a single prediction. To address multi-modality, NSM is enhanced with a Conditional Variational Autoencoder (CVAE), allowing for multiple transform predictions, with the top-scoring transform selected for execution. ``NSM-base" and ``NSM-base + CVAE" are considered as two different baselines.

\noindent \textbf{Relational Pose Diffusion (RPDiff):} RPDiff operates directly on 3D point clouds and is capable of generalizing across novel geometries, poses, and layouts. It addresses the challenge of multiple similar rearrangement solutions through an iterative pose de-noising training strategy, allowing for precise, multi-modal outputs. It was the state of the art method on RPDiff's benchmark until the present work.
% By focusing on local geometric features and disregarding irrelevant global structures, RPDiff ensures enhanced generalization and accuracy across various rearrangement tasks in both simulation and real-world scenarios.

% \noindent\includegraphics[width=\linewidth]{example-image} 

\begin{table}[]
\centering
\begin{tabular}{@{}lcc@{}}
\toprule
\textbf{Method}     & \textbf{Book/Shelf} & \textbf{Can/Cabinet} \\ \midrule
C2F Q-attn          & 57$\%$                & 51$\%$                 \\
R-NDF-base          & 00$\%$                & 14$\%$                 \\
NSM-base            & 02$\%$                & 08$\%$                 \\
NSM-base + CVAE     & 17$\%$                & 19$\%$                 \\
RPDiff              & 94$\%$                & 85$\%$                 \\
\textbf{DAP (ours)} & \textbf{98}$\%$                    &  \textbf{94}$\%$                    \\ \bottomrule
\end{tabular}
\caption{Performance on RPDiff benchmark (Success rate).} \label{tab:performance}
\vspace{-2em}
\end{table}

The comparative performance of DAP and the baselines is presented in TABLE~\ref{tab:performance}. The table illustrates that RPDiff significantly outperforms the other four baselines. However, DAP exceeds RPDiff's performance by a considerable margin, highlighting DAP's superior capability. Furthermore, the efficiency of DAP is demonstrated through its training requirements: RPDiff necessitates three days of training on a V100 GPU for its action module and an additional five days for the evaluation module on each task. In contrast, DAP requires only one hour for training the affordance prediction module and another hour for pose estimation, all on a single 3090 GPU, showcasing DAP's remarkable efficiency.

\begin{table}[]
\centering
\begin{tabular}{@{}lcc@{}}
\toprule
\textbf{Method}     & \textbf{Book/Shelf} & \textbf{Can/Cabinet} \\ \midrule
CAP          &  24$\%$      &     36$\%$   \\ 
\textbf{DAP (ours)} & \textbf{98}$\%$                    &  \textbf{94}$\%$           \\ \bottomrule
\end{tabular}
\caption{Ablation study on RPDiff benchmark.} \label{tab:ablation}
\end{table}

\subsection{Ablation Study}
To analyze the impact of diffusion-based affordance prediction, we conducted an ablation study upon RPDiff benchmark. We compare DAP with a variant framework without using DDPM loss:

\noindent \textbf{Classification Affordance Prediction (CAP):} Instead of treating affordance prediction as a generative task, we approach it as a classification problem using cross-entropy loss. The architecture remains the same as DAP. During inference we do not perform iterative de-noising, but provide the classification in one step.

% \noindent \textbf{Direct Pose Estimation (DPE):} Instead of utilizing correspondence prediction combined with Arun's algorithm for pose determination, we examine a variant of DAP that directly outputs pose estimation. We adapt our architecture, as shown in Fig.~\ref{fig:corr}, by eliminating the correspondence matrix prediction and incorporating a pose decoder. This decoder predicts the translation vector and rotation axis, which previous work~\cite{sundermeyer2023bop} has shown to be effective and stable.

The results of our ablation study are depicted in TABLE~\ref{tab:ablation}. Classification Affordance Prediction (CAP) significantly underperforms compared to the complete DAP. This finding confirms our hypothesis that diffusion-based affordance prediction effectively addresses multi-modality issues.

\subsection{Real world Experiment}

\begin{figure}
    \centering
    \includegraphics[width=0.7\linewidth]{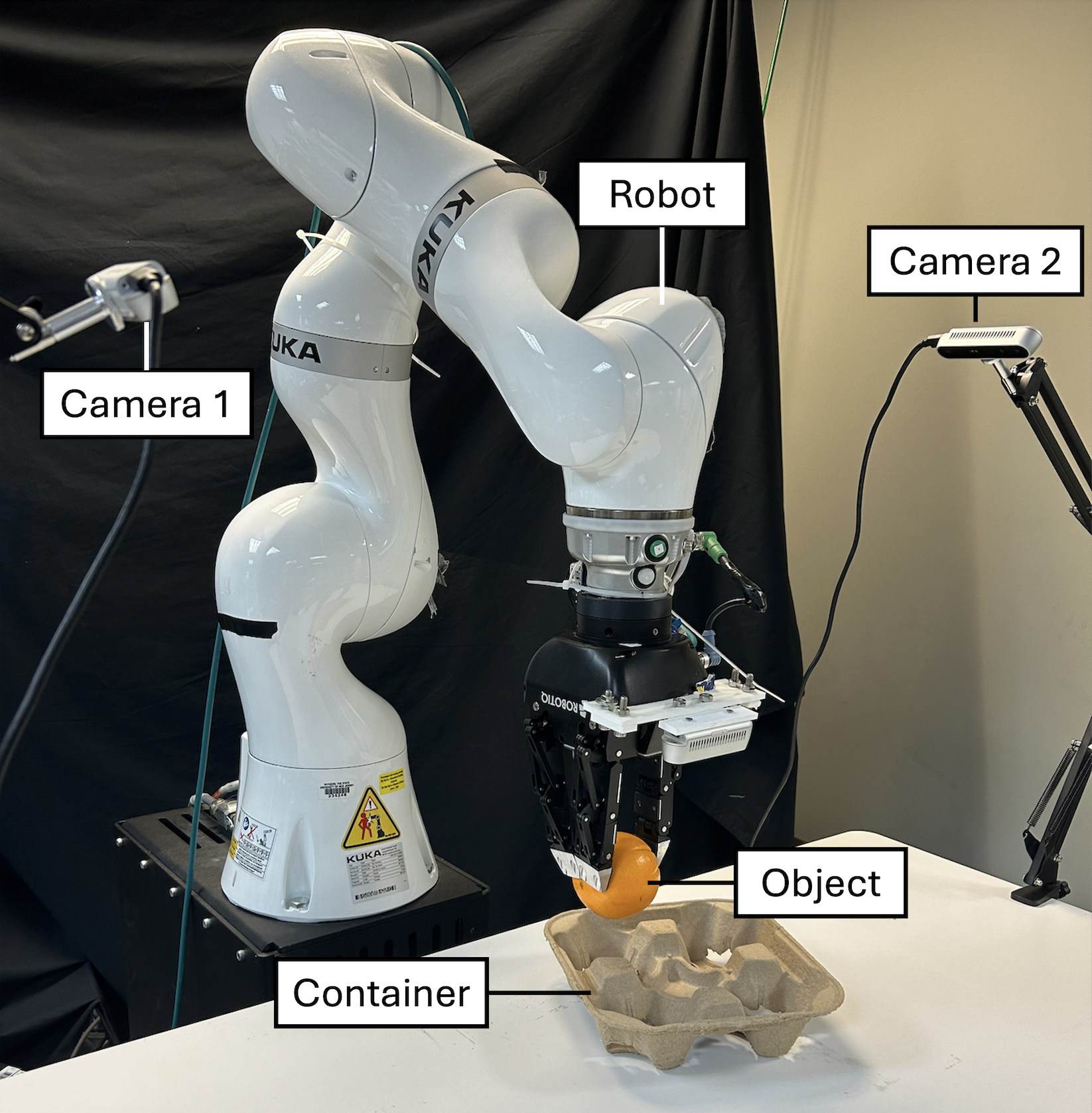}
    \caption{\small{Robot setup: a Kuka robot equipped with two RealSense D415 cameras and a three-finger Robotiq hand. }}
    \label{fig:robot-setup}
    \vspace{-1em}
\end{figure}

\begin{figure*}
    \vspace{-2em}
    \centering
    \includegraphics[width=0.9\linewidth]{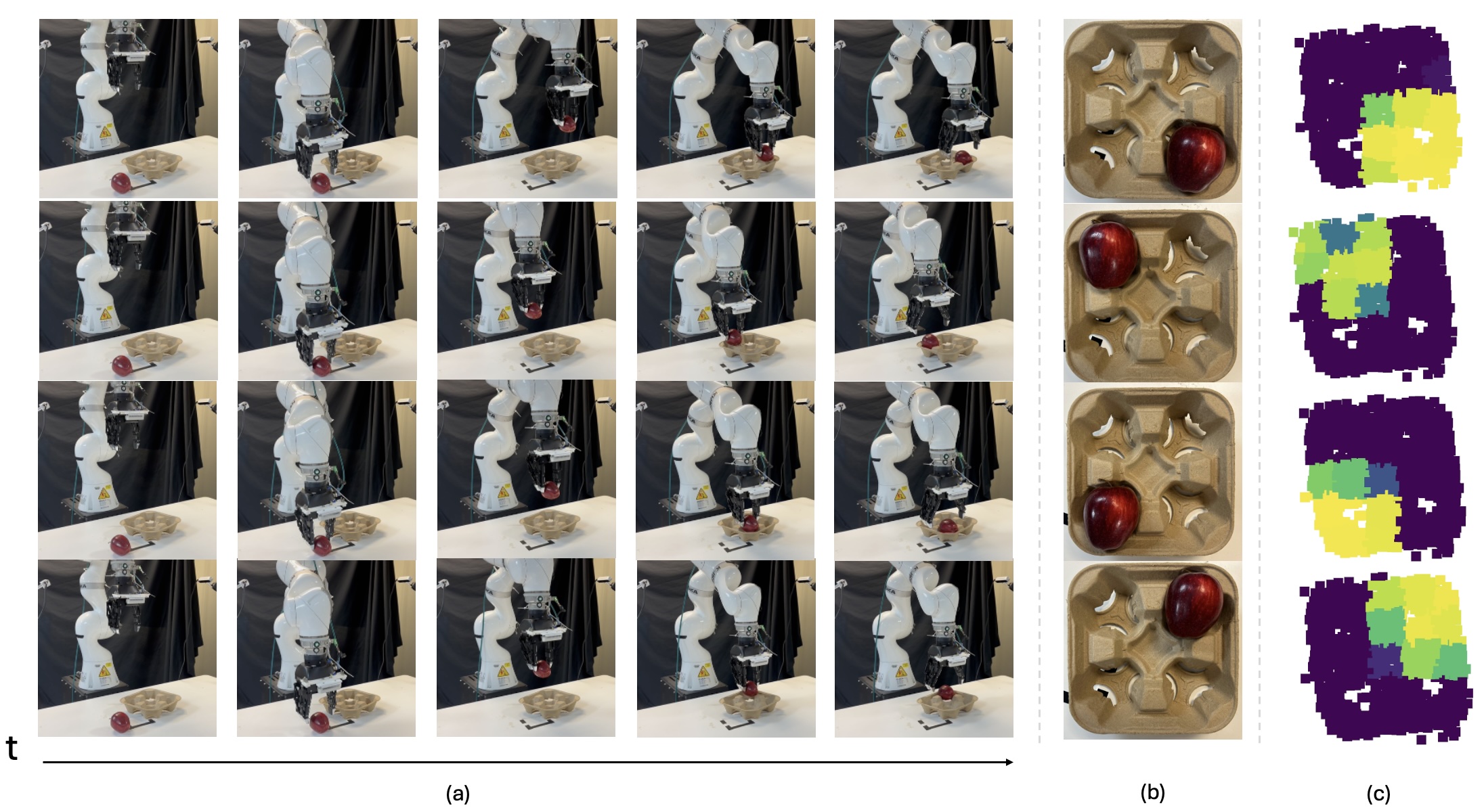}
    \vspace{-0.5em}
    \caption{\small{Real experiment on fruit storage. From left to right are: (a) robot execution recording; (b) final storage result; and (c) placeable affordance prediction.}}
    \label{fig:real_exp}
    \vspace{-1em}
\end{figure*}

\begin{figure}
    \centering
    \includegraphics[width=0.7\linewidth]{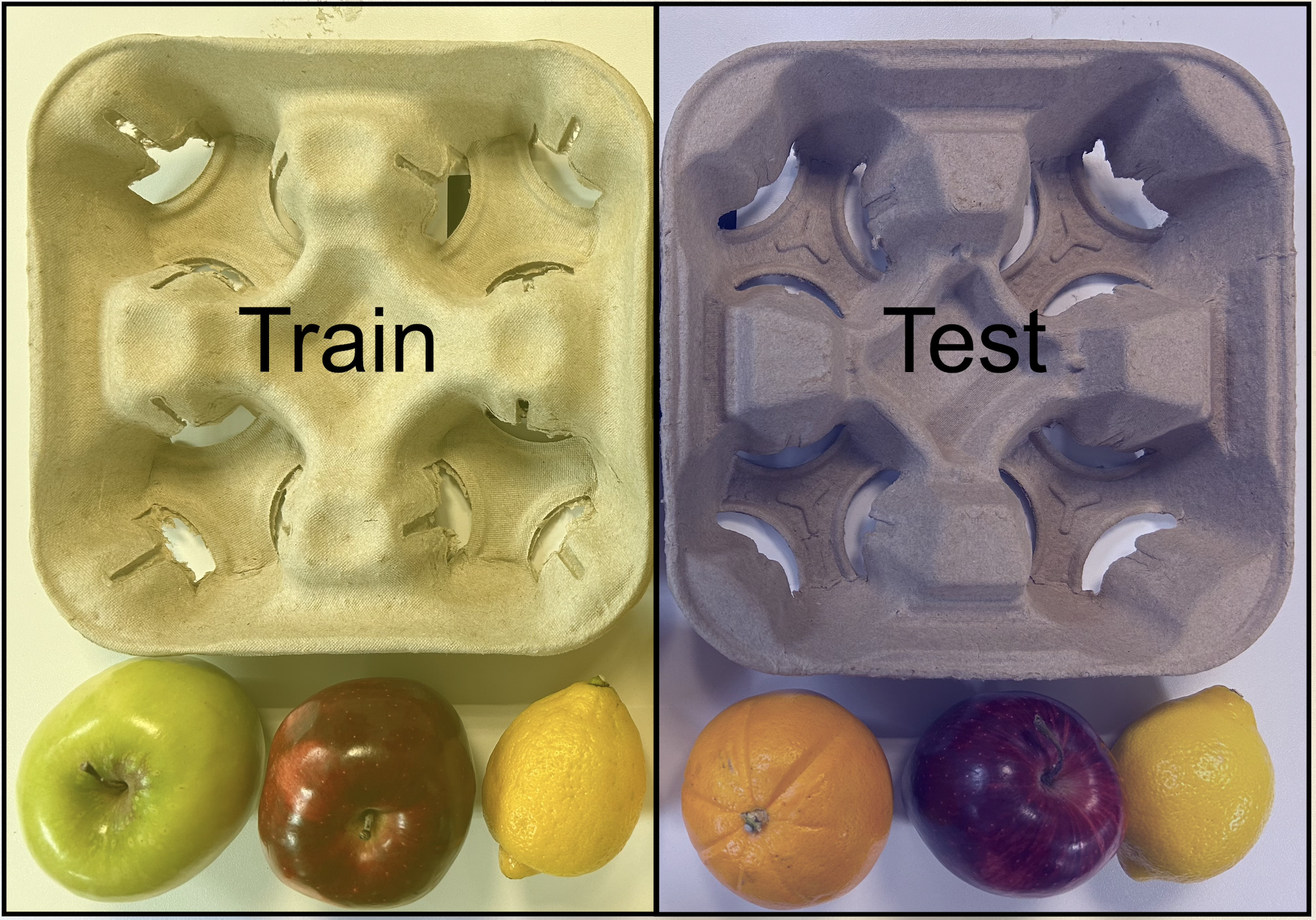}
    \caption{\small{Task objects: the fruits and storage racks used in the real-world fruit storage task for training (left) and testing (right).}}
    \label{fig:robot-objects}
    \vspace{-2em}
\end{figure}

We conducted a qualitative real-world experiment to assess DAP's performance in a real-world setting, using a real-to-real setup for both data collection and deployment. This approach, distinct from the sim-to-sim or sim-to-real setups in previous works \cite{shelving,liu2022structdiffusion}, faces challenges from noisier data and a significantly smaller dataset. Unlike the thousands of clean data points available from simulations, real-world data is inherently noisier and scarcer. To our knowledge, DAP is the first to demonstrate real-to-real capabilities in tackling the multi-modality storage problem, a notable advancement over prior diffusion-based methods like RPDiff\cite{shelving} and StructDiffusion~\cite{liu2022structdiffusion}, which have leaned on sim-to-real setups. Existing imitation-based rearrangement frameworks such as Transporter networks~\cite{zeng2021transporter} and CLIPort~\cite{shridhar2022cliport} can deal with real-to-real setup but fall short in addressing multi-modality issues. Addressing the real-to-real multi-modality storage problem necessitates a delicate balance between the model's representational capacity and data efficiency.

\noindent \textbf{Robot Setup:} We conducted real-world experiments using a Kuka IIWA 14 robot arm, equipped with a Robotiq 3-finger adaptive gripper. We positioned two Intel RealSense D415 cameras on opposite sides to observe the container and the object at the same time. This setup is illustrated in Fig.~\ref{fig:robot-setup}.

\noindent \textbf{Task:} We trained and tested our method in a real-world fruit storage task, where given the initial point clouds of a fruit and a storage rack, the robot arm is asked to pick and place the fruit into one of the four rack slots (Fig.~\ref{fig:robot-objects}). We collected 80 demonstrations. The fruits and storage rack for the testing experiments are unseen during training. Each demonstration consists of a start scene point cloud and an end scene point cloud. We used the Segment-Any-Thing (SAM)~\cite{kirillov2023segment} model to segment out the fruits and storage rack.
The testing demos in Fig. ~\ref{fig:real_exp} show that our method can generalize well to unseen objects and containers in the real world. As shown in Fig~\ref{fig:real_exp}, DAP successfully detected all four placeable regions on the test storage rack.

\section{Limitations}
There are several limitations in the current DAP pipeline. (1) DAP is primarily limited to storage problems, wherein a target object is placed inside a larger container. In this setup, after applying the diffusion-based affordance segmentation to the container's point cloud, there is only one optimal goal pose for the local region. For tasks where multiple optimal solutions exist within the local region, such as hanging a cup, which can be hung by its handle or by its rim, DAP performs less effectively. Investigating how to combine diffusion models with correspondence prediction can be a future research direction. Moreover, 
the storage of arbitrary target objects necessitates the use of open-set object detectors~\cite{zhang2023optical, zhang2023detect, zhong2022regionclip}.
(2) A multi-camera setting is required for real-world applications, as it helps maintain consistency between the point cloud data during training and deployment. This limitation could be addressed through advancements in point cloud backbones or by developing new 3D data augmentation techniques.
\section{CONCLUSION}
We present DAP, a diffusion-based affordance prediction pipeline for multi-modality storage problems, aimed at placing a target object into a larger container. Our method consists of two steps: a diffusion-based prediction step and a pose estimation step. First, we sample a placeable region for the container using a diffusion model and crop it out. Then, we compute the point-wise correspondence between the target object and the cropped region of the container. This correspondence indicates which parts of the two geometries should be in contact. We employ Arun's algorithm to solve the goal relative pose of the object with respect to the container. Through thorough experimentation, including both simulation and real-world scenarios, we demonstrate that our proposed DAP pipeline is superior in performance and training efficiency compared to previous methods. We hope that DAP can pave the way for further research on multi-modality pair-wise object manipulation tasks.
placeholder

\bibliographystyle{IEEEtran}
\bibliography{Star} % Entries are in the Star.bib file
\appendix
\subsection{Implementation Details}
Similar to Mask3D~\cite{mask3d} and Oneformer3D~\cite{oneformer3d}, the point cloud $\mathbf{P}$ in this work has been pre-clustered using the super-point algorithm~\cite{sun2023superpoint}. Each point in $\mathbf{P}$ represents a super-point rather than a raw point. The position $v$ and the normal $n$ represent the average position and normals of all raw points within the cluster of the superpoint.
.  % Comment out for IROS final version
\end{document}